# Membrane-based Acoustic Microrobots

Fatih Kocabas[1], *Student Member, IEEE,* Cemal Polat Avdar[1], Prithvi S Venkatesh[1], and Yunus Alapan[1,3,4], *Member, IEEE*

***Abstract*— Acoustic microrobots have emerged as a promising frontier for targeted drug delivery and minimally invasive medicine due to their high-power density and biocompatibility. Despite wide-ranging designs, conventional acoustic microrobots mostly rely on air microbubbles trapped within confined microcavities within the robot body, which suffer from limited operational longevity due to rapid gas dissolution and resultant shifts in resonance frequency. In this paper, we propose a robust, membrane-based acoustic microrobot that overcomes these limitations by employing a thin flexible Polydimethylsiloxane (PDMS) membrane bonded over confined microcavities for microstreaming. The introduced design physically prevents gas diffusion, ensuring stable performance over extended periods at high actuation voltages. We systematically characterized the membrane-based acoustic actuator longevity, demonstrating consistent streaming and propulsion for over 24 hours of continuous operation. In addition, by embedding magnetic microparticles into the structural body, these actuators were successfully employed as microswimmers with directional control using low-intensity (2 mT) external magnetic fields. Finally, we demonstrate the scalability of the proposed design architecture down to ~100 μm. This membrane-based approach establishes a reliable framework for the development of high-endurance acoustic microactuators and microrobots capable of performing long-term tasks.**

## I. INTRODUCTION

Wireless microrobots have gained significant interest in the past decade due to their unique capabilities in navigation and manipulation at small scales with wide-ranging applications in minimally invasive surgery, cell manipulation, and targeted drug delivery [1]–[9]. Among different methods of actuation for microrobots, including optical, electrical, and magnetic, acoustic actuation stands out owing to the ability to apply localized fields over longer distances and being safe to penetrate through biological tissues [10], [11].

Acoustic microrobots are typically constructed from polymeric materials featuring engineered microcavities to trap microbubbles [2], [12]–[14]. Exposing these robots to an acoustic field induces rapid microbubble oscillation, generating microstreaming [15], [16]. Consequently, particularly at resonant frequencies, the microrobot experiences directed propulsive forces that enable navigation [17]. The ability to generate thrust via microbubble oscillation has enabled the development of mobile acoustic microactuators and microrobots, including targeted microswimmers [14], [18], [19], precision micromanipulators [1], and artificial muscles [20]. However, recently proposed methods report issues with long-term robustness due to a gradual loss of propulsive power. This degradation occurs because the trapped gas continuously diffuses into the surrounding fluid, leading to natural frequency changes during operation and, ultimately, the complete dissolution of the microbubble [14], [21].

In the literature, several approaches have been pursued to tackle this challenge, ranging from chemical functionalization approaches to engineered microcavities. Rendering microcavities hydrophobic through fluorosilane-based coatings enable enhanced anchoring of microbubbles within microcavities and increase their lifetime [12], [22]. In addition, engineered microcavities with carefully tuned openings smaller than the trapped bubble can effectively control gas diffusion interface and slow down the bubble degradation [14], [21]. However, both approaches require complex surface functionalization or fabrication processes, respectively, while the increase in robustness remains minimal. In the literature, the microstreaming generated by thin membranes under external acoustic field exposure has been previously theorized [23], [24]. Recently, different groups have experimentally demonstrated microstreaming generating capabilities of membrane-based systems, though at the millimeter scale [20], [25]. Therefore, it remains to be explored whether this approach can be effectively scaled down and deployed for the propulsion of microrobots.

In this paper, we propose an acoustic microrobot design sealing the internal microcavities with a thin, flexible membrane to overcome the fundamental limitations of bubble dissolution. Thin membranes physically prevent gas diffusion into the surrounding fluid and undergoes rapid oscillation under acoustic excitation, generating the necessary microstreaming for propulsion (Figure 1). Consequently, these microrobots achieve longer operation times while maintaining robust and highly repeatable performance. The details of manufacturing of the acoustic microrobot and the actuation setups is explained in Section II. The systematic characterizations for the longevity and voltage-velocity relationship, along with a membrane-based magneto-acoustic microrobot, followed by a scaling down demonstration is given in Section III. The concluding comments remarking possible future enhancements can be found in Section V.

[1] F. Kocabas, C. P. Avdar, P. S. Venkatesh, and Y. Alapan are with the Bio- integrated Robotics Lab at Mechanical Engineering Department, University of Wisconsin - Madison, Madison, WI 53706 USA (608-265-5673; alapan@wisc.edu).

[2] Y. Alapan is with the Biomedical Engineering Department, University of Wisconsin-Madison, Madison, WI 53706 USA.

[3] Y. Alapan is with Carbone Cancer Center, University of Wisconsin-Madison, Madison, WI 53706 USA.

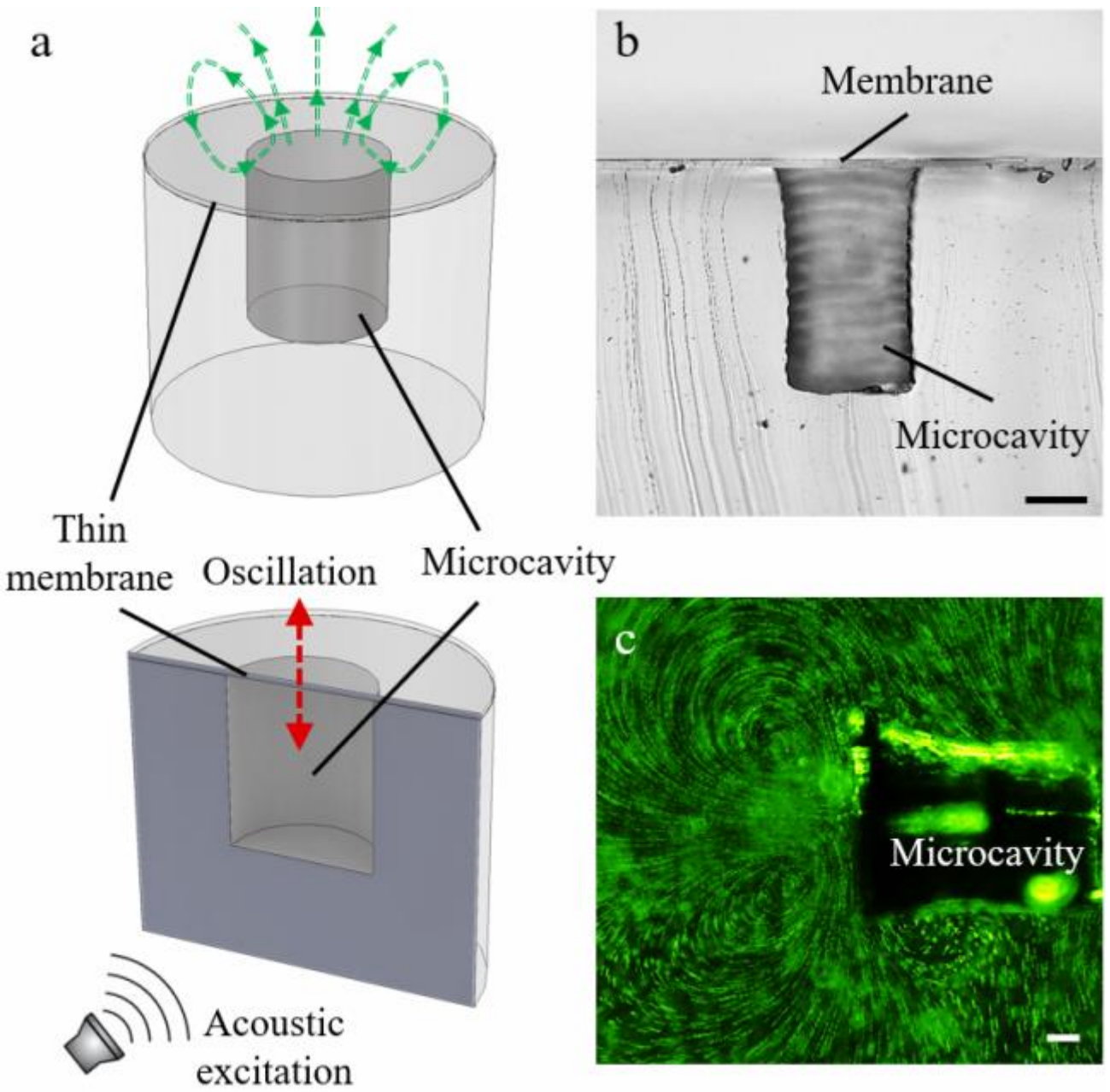

Figure 1. Membrane-based acoustic microrobots. a) Microstructures with confined microcavities and overlaid with a thin membrane allow acoustic excitation and microstreaming. b) Bright field microscopy image of a membrane-based microrobot cross-section. c) Microstreaming tracing particles patterns generated by a membrane-based acoustic microrobot captured under fluorescent microscope. Scale bars, 200 µm.

## II. MANUFACTURING

### A. The Acoustic Microrobot

The fabrication strategy for the acoustic microrobot body utilized high-resolution 3D printing and soft lithography. Molds featuring 350 µm diameter x 500 µm length pillars of 3x3 arrays were produced using a lithography-based printer (Form 4, FormLabs) with a commercial resin (Gray v5, FormLabs). Post-printing, the molds were rinsed in isopropyl alcohol (IPA) to remove residual uncured monomer, followed by a one-hour UV post-cure and a two-hour thermal treatment at 120°C. To ensure complete removal of resin traces and guarantee non-inhibition of PDMS curing, the molds were submerged in an IPA bath overnight. Prior to casting, the molds were functionalized with Trichloro(1H,1H,2H,2H-perfluorooctyl)silane (MilliporeSigma) via vapor-phase deposition to facilitate demolding. PDMS (Sylgard 184, Dow Inc.) was prepared at a 10:1 (base:curing agent) weight ratio, degassed, and cast into the molds. Entrapped air bubbles around the micropillars were manually removed using fine-tip tweezers to ensure structural uniformity. The assembly was cured on a hotplate at 80°C. Post-curing, the PDMS blocks were demolded and sequentially sonicated in IPA and DI water for 30 minutes each. The microrobot membrane was fabricated by spin-coating PDMS onto a silanized glass substrate at 2000 rpm for 60 seconds, yielding a thickness of 25 µm. Both the PDMS block and the membrane underwent oxygen plasma treatment to enable covalent bonding. Finally, the bonded structures were cut out using a razor blade to achieve their final geometry (Figure 2).

To provide the microrobots with directional steering and navigation capabilities, magnetic particles were integrated into the structural design. This functionalization allows for precise control via external magnetic fields of approximately 2 mT. The fabrication utilized the previously described 3D-printed molds as a template. Initially, a base layer of pure PDMS was cast until the micropillars were fully submerged and subsequently cured on a hotplate. A secondary composite layer, comprising PDMS mixed with NdFeB magnetic particles (Magnequench, 20065-089) at a 1:2 mass ratio, was then casted on top, and cured. This manufacturing method is developed to ensure that plasma treatment is not blocked by the attraction of magnetic particles. The resulting composite PDMS block was bonded to a 25 µm thick, spin-coated PDMS membrane following ozone surface treatment. After debonding from the glass substrate, the membrane-based microrobots were cut out using a 1 mm diameter biopsy punch to define their final geometry. To finalize the magnetic functionality, the robots are axially magnetized by high-intensity magnetic field (~1.1 T) using an N52 permanent magnet (1” diameter x 1” length, K&J Magnetics).

To fabricate membrane-based acoustic microrobots at the 100 µm scale, a modified mask photolithography process was employed. A physical reservoir was created by bonding laser-cut double-sided tape (8146, 3M) with a rectangular aperture onto a glass slide. This cavity was filled with an epoxy-based negative photoresist (SU-8, Kayaku Advanced Materials). To ensure a stable soft-bake, the resist was heated to 95°C at a ramp rate of 5°C/min, held for two hours, and subsequently cooled to room temperature at the same rate (5°C/min) to preserve structural integrity and minimize internal stress. UV exposure was performed for 8 seconds using a high-intensity curing spot light (Agiltron), integrated with a diffuser to ensure irradiance homogeneity across the resist surface. Following exposure, the samples were developed in PGMEA (MilliporeSigma) until the uncrosslinked regions were fully removed. A final post-exposure bake (PEB) was conducted at 95°C with controlled temperature ramping to finalize the polymerization. The resulting SU-8 master mold was functionalized with a silane coating to facilitate subsequent replication. PDMS was cast over the mold and allowed to cure at room temperature to minimize thermal shrinkage. Following demolding, the PDMS layer—featuring the negative microcavity structures—was bonded via oxygen plasma treatment to a thin PDMS membrane. This membrane was prepared by spin-coating at 8000 rpm for 60 seconds, resulting in a thickness of approximately 8 μm.

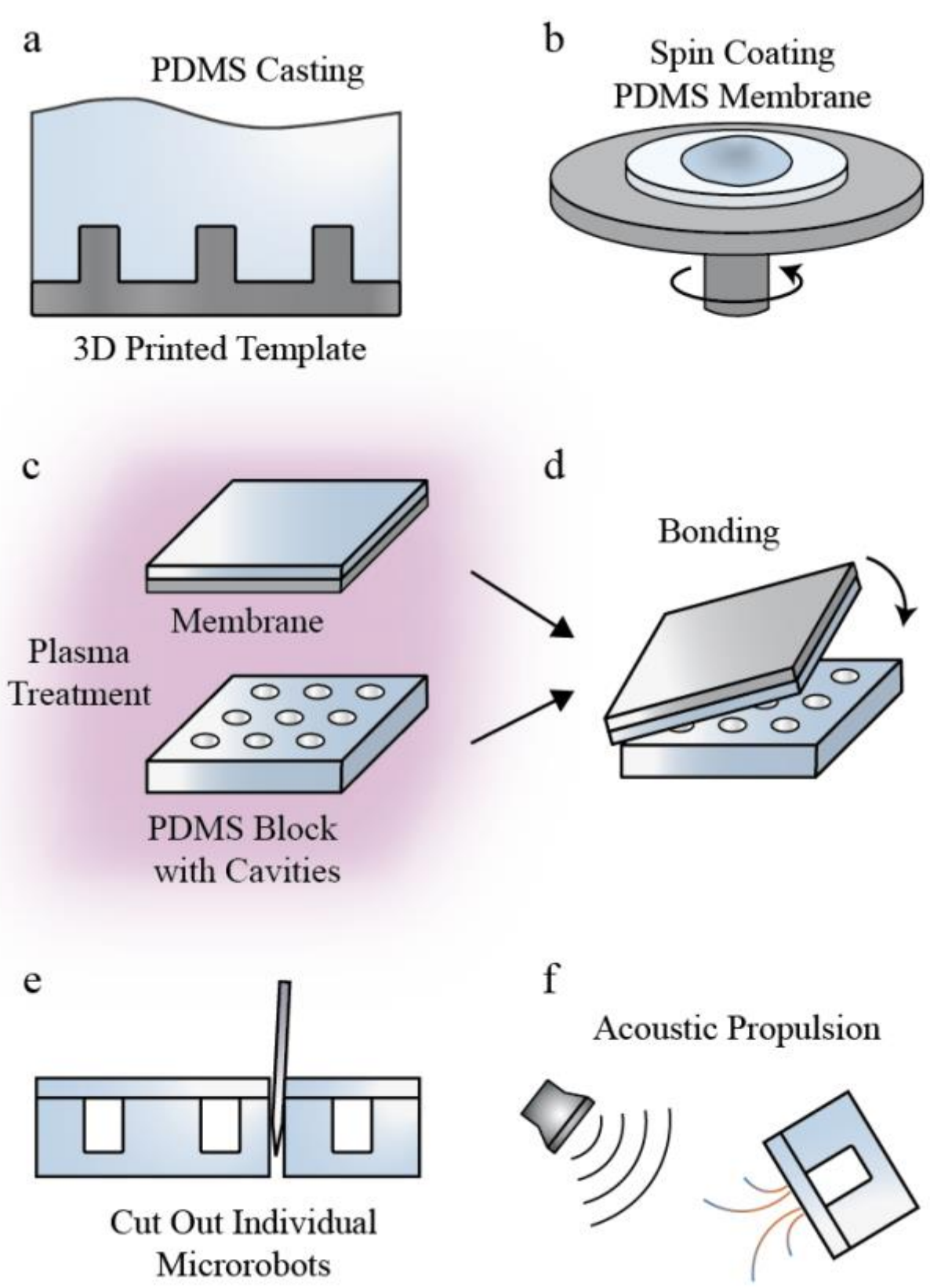

Figure 2. Fabrication of membrane-based acoustic microrobots. a) PDMS casting over 3D printed templates with micropillars results in microcavities. b) PDMS spin coated at specific speeds to generate thin membranes. c, d) Plasma treatment of fabricated thin membrane and PDMS block with microcavities (c) enable their bonding to each other (d). e, f) Careful cut outs of structures with individual or multiple microcavities (e) allow actuation of microrobotic structures under acoustic fields.

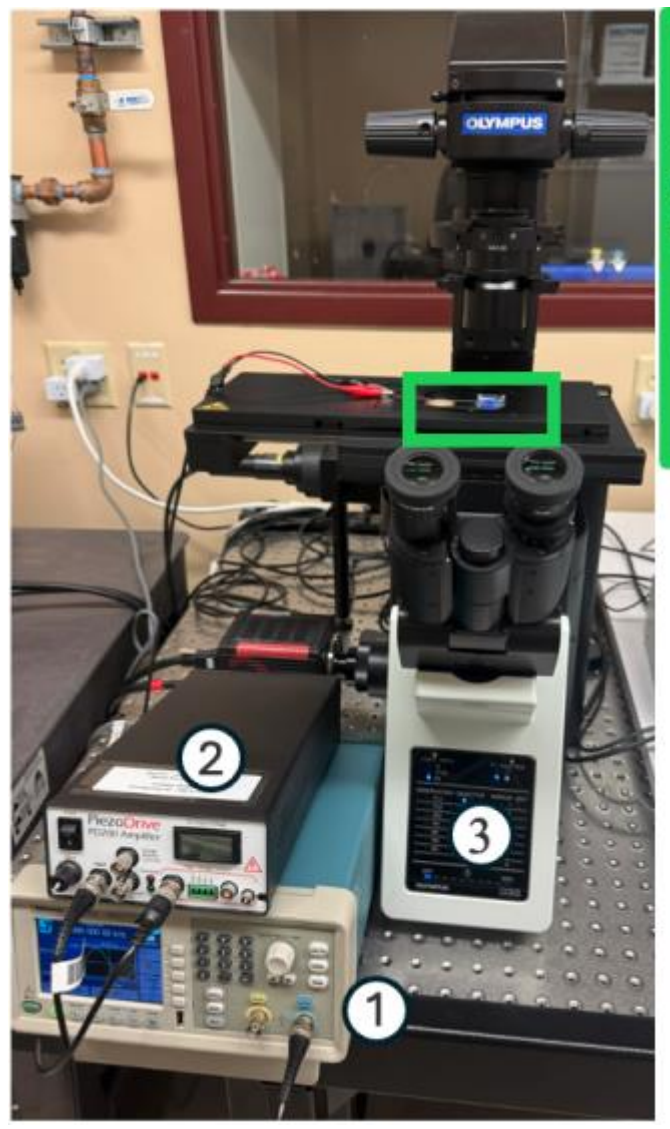

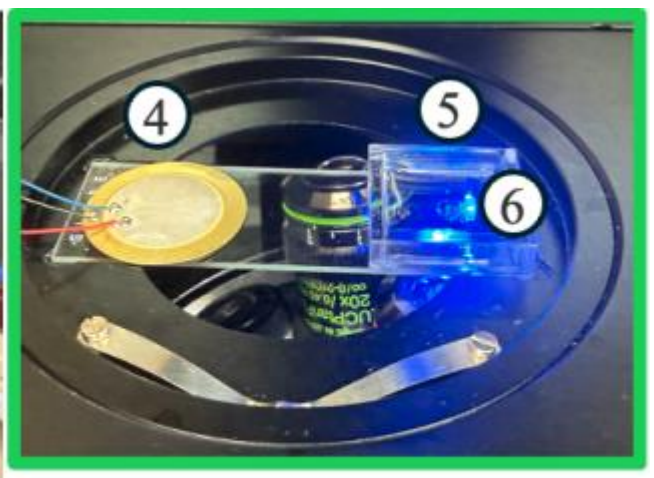

Figure 3. The setup used for acoustic actuation of membrane-based microactuators and their performance characterization. Inset shows close-up view of piezo transduced along with actuation chamber on an inverted microscope stage.

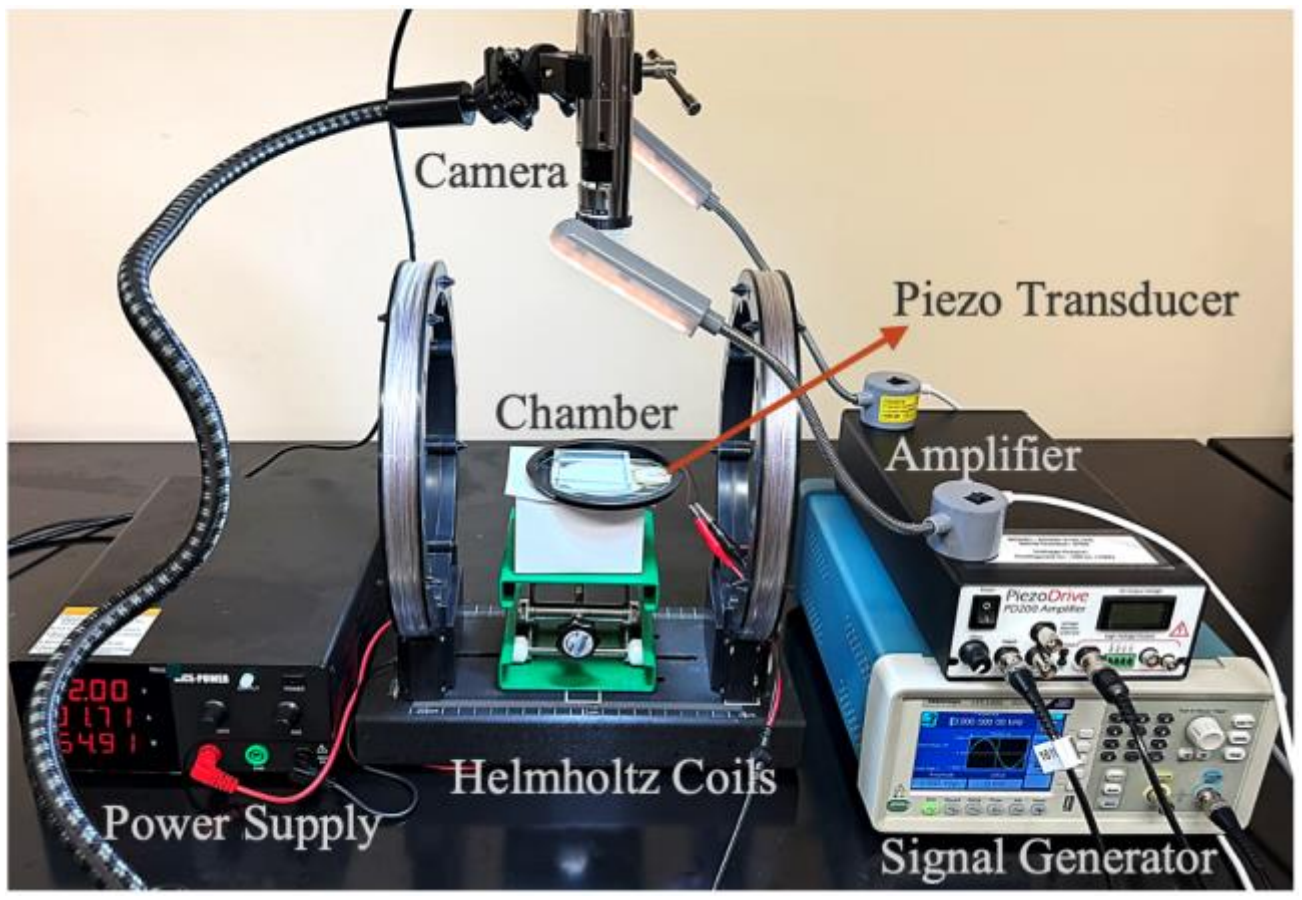

Figure 4. The setup used to control the membrane-based magneto-acoustic microrobots.

### *B. Acoustic Actuation Setups*

To characterize the fluid flow and acoustic actuation, an experimental setup was developed utilizing a piezoelectric transducer (7BB-27-4CL0, Murata Electronics) as the acoustic source. The transducer was bonded to one side of a glass slide using a epoxy adhesive. Signal generation was provided by a function generator (AFG1022, Tektronix) and subsequently amplified through a power amplifier (PD200, Micromechatronics Inc.).

The water chamber was constructed using an acrylic sheet, pre-laminated with double-sided adhesive tape (8146-3, 3M), and patterned via laser cutting to define the chamber walls. This assembly was adhered to the glass slide, opposite the transducer. For flow characterizations, a 3” × 1” glass slide was used to support a 1” × 1” chamber. The pool was filled with Phosphate-Buffered Saline (PBS) mixed with with 2 μm diameter fluorescent tracer particles (L4530, Sigma-Aldrich) at a 0.2% volume ratio (Figure 3).

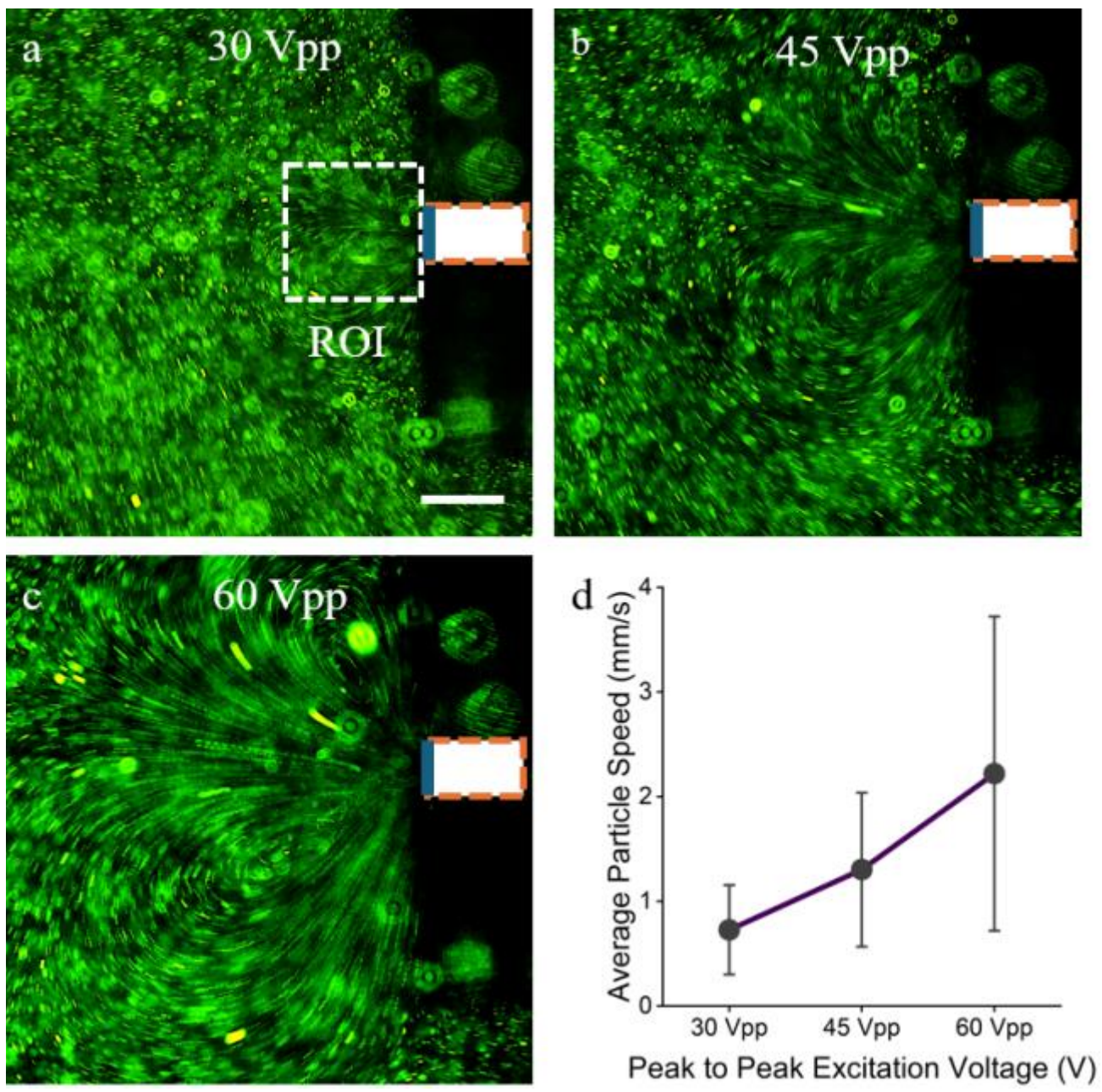

Figure 6. Effect of excitation voltage on microstreaming speed of membrane-based acoustic microactuators. Microstreaming patterns at a) 30 Vpp, b) 45 Vpp, c) 60 Vpp. d) Average tracer particle speed vs voltage measurements show increasing microstreaming speed at greater voltages and error bars represent standard deviation (n = 3). Scale bar, 500 μm.

For navigation demonstrations, a 3” × 2” glass slide was utilized with a 2” × 2” chamber. The chamber was subsequently filled with PBS mixed with 0.1% Tween 20 (MilliporeSigma) to reduce surface tension. The acoustic excitation setup was positioned at the center of a pair of Helmholtz coils consisting of 500 turns of 22 AWG enameled copper wire per coil. The coils, featuring ~10 cm inner and ~11.4 cm outer radii, were supplied with 32 V to generate a homogeneous magnetic field with a magnitude of about 2 mT.

The motion of the membrane-based acoustic microrobot was recorded using a commercially available digital microscope (EdgePlus, Dino-Lite). The piezo transducer, signal generator, and amplifier were identical to the actuation setup used for flow characterization (Figure 4).

## III. Experiments & Results

### A. Longevity Experiments

Longevity tests were performed to evaluate the performance over time. A PDMS block which was bonded to a 25 µm PDMS membrane was cut with a razor to contain two membrane-based acoustic microrobot units and bonded to the base glass of the water chamber after plasma treatment of both surfaces. After bonding, the chamber containing the PDMS block with membraned microcavities, plasma treated again to diminish the bubble formation at undesired locations within chamber. The empty water chamber with the bonded PDMS block to the glass base, is later filled with 2 µm diameter tracer particles mixture with phosphate-buffered saline (PBS) solution at a 0.2% volume ratio. The acoustic setup containing the PDMS block was placed on an inverted fluorescent microscope (Olympus IX-83) to image the generated flow by the proposed method. The natural frequency of the membranes was found by sweeping the excitation frequency of the piezo transducer while observing the generated microstreaming. This process resulted in a resonant frequency of 13 kHz. The piezo transducer was actuated with 60 volts peak to peak with sinusoidal waveform, at natural frequency of membrane. Videos, later to be used for performance characterization, were recorded at the beginning and after 1, 3, 5, 6, and 24 hours of constant actuation. For the first 6 hours, as water evaporated, water was added manually; before leaving the system overnight for the 24-hour longevity test, we sealed the ceiling of water chamber with double-sided tape to ensure the water level remained stable. During the experiments, a slight downward shift of about 0.2 kHz in frequency over time was observed, and the excitation frequency of piezo transducer was adjusted to make sure the system was actuated at the natural frequency.

To quantify the generated microstreaming, the motion of the 2 µm tracer particles was analyzed using a custom-developed particle tracking velocimetry (PTV) script. The software was implemented to process the recorded microscope videos through a sequential image processing pipeline. First, the frames were converted to grayscale and subjected to a background subtraction filter to eliminate static noise and enhance the contrast of the moving particles.

Following pre-processing, a Gaussian blur was applied to reduce high-frequency electronic noise, and a thresholding operation was used to binarize the frames. The centroids of individual particles were identified in each frame and linked across consecutive time steps using a nearest-neighbor search algorithm with a defined maximum displacement limit. This allowed for the reconstruction of individual particle trajectories. The local velocity vectors were calculated by dividing the spatial displacement by the known frame rate, which was 4.822 ms, of the microscope. Finally, the tracking data was used to calculate the average speed of tracer particles, within the same region of interest for each sample.

The robustness of performance, quantified from average particle speed over time is proving that the proposed method can generate consistent microstreaming even after 24 hours of continuous actuation (Figure 5e).

### B. Voltage vs Velocity

For the acoustic streaming experiments, the PDMS block containing membrane-based actuation units are fixed on glass slide using plasma treatment, followed by a plasma treatment of to be water exposed surfaces to diminish unwanted bubble formation. 2 µm diameter tracer particles were suspended in a phosphate-buffered saline (PBS) solution at a 0.2% concentration ratio and poured into the water chamber. An initial frequency sweep from 1 kHz to 200 kHz with 60 volts peak to peak with sinusoidal waveform is utilized for actuating the piezo transducer, revealed the system's resonant frequency at 13 kHz, which was utilized for all subsequent experiments. The generated acoustic fields were then evaluated at varying excitation voltages of 30, 45, and 60 volts peak to peak sinusoidal waveform supplied to the piezo transducer. Using a custom-written tracking algorithm, identical to the one used for longevity experiments, particle velocities were measured specifically within a defined region of interest (Figure 6a). To ensure reliability of the characterization procedure, the voltage-velocity data was collected and averaged across three independent samples. The resulting plot (Figure 6d) demonstrates that average particle speed increases with higher excitation voltages, indicating a corresponding increase in the generated thrust.

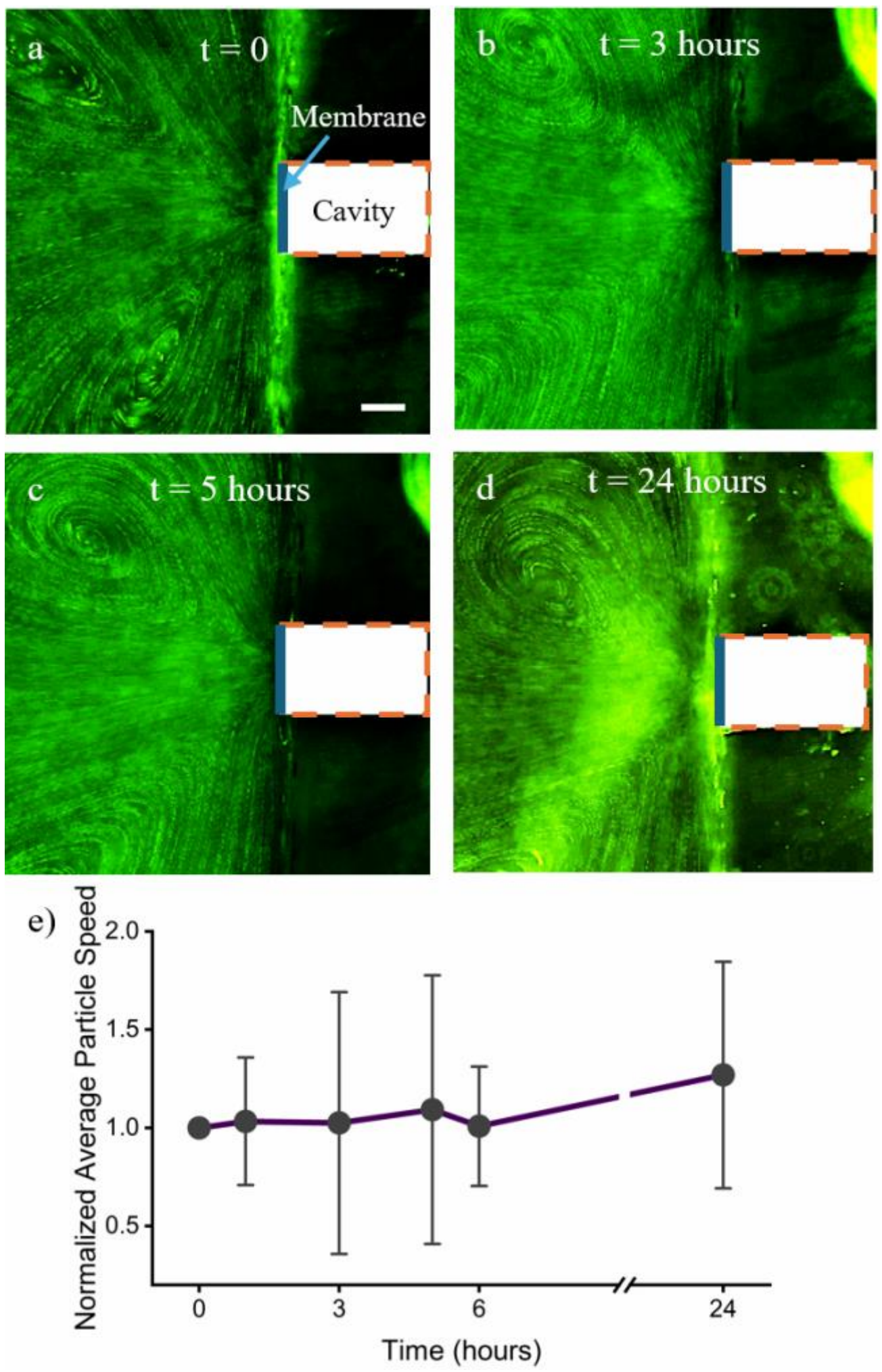

Figure 5. Longevity and robustness evaluation of the membrane-based acoustic microactuators. a) Microstreaming patterns generated at the start of the experiment. b-d) Microstreaming profile after 3 (b), 5 (c), and 24 (d) hours of continuous actuation. e) Normalized particle speed vs time (hours) results show robust performance over 24 hours. Data points show mean of tested samples and error bars represent standard deviation. (n = 4). Scale bar, 200 µm.

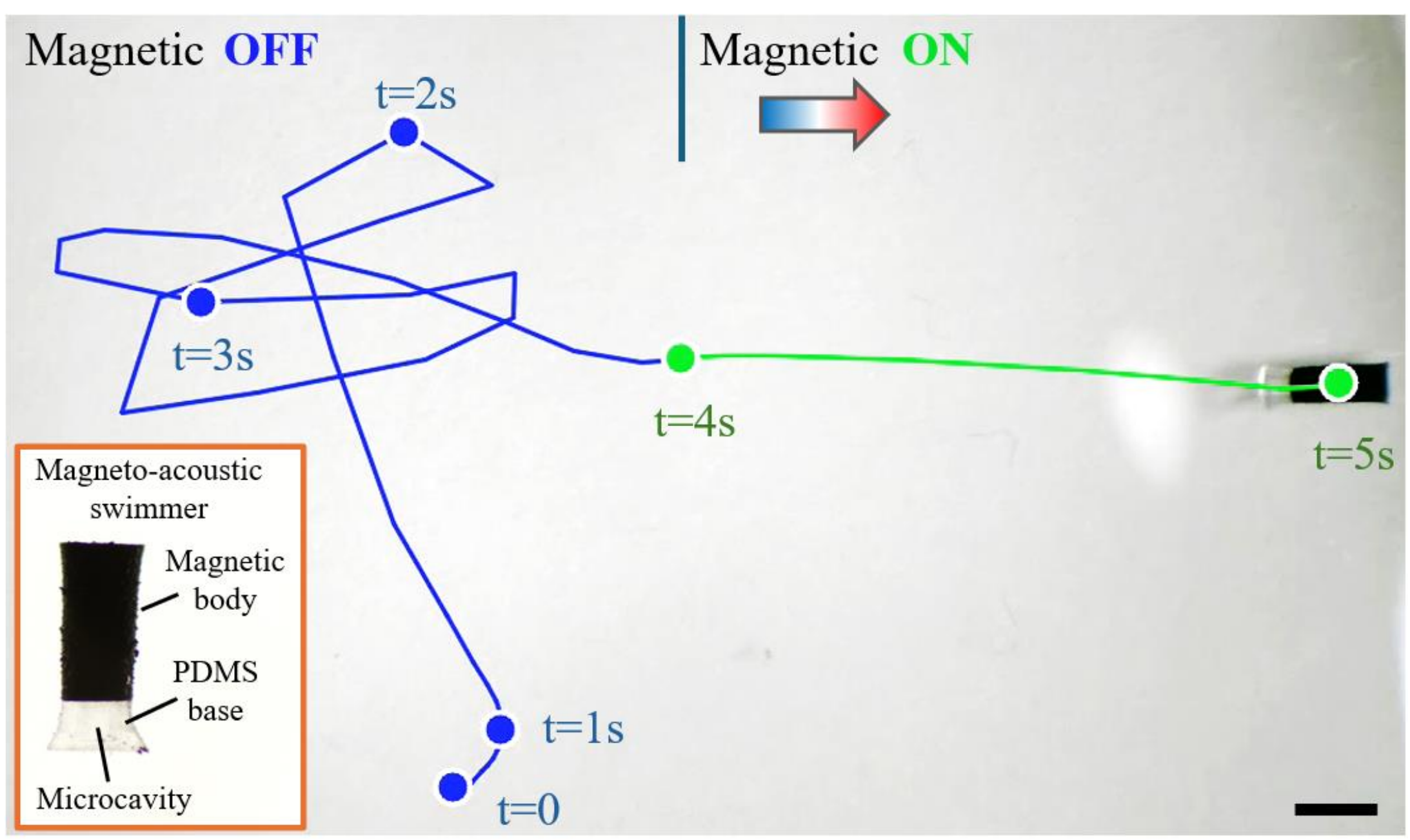

Figure 7. Membrane-based acoustic microrobot actuation: Under only acoustic field, resulting in random path following (left, blue line). Under both acoustic and magnetic fields, following a predetermined path dictated by homogenous external magnetic field. Scale bar 1 mm.

*C. Membrane-based Magneto-Acoustic Microrobot*

To demonstrate the enhanced capabilities of the platform, a hybrid membrane-based magneto-acoustic microrobot was developed (Figure 7). The fabrication utilized a sequential two-step casting process: initially, a base layer of pure PDMS was cast into the mold to define the functional microstructures; following curing, a secondary layer of PDMS doped with magnetic particles was cast to form the magnetic steering component. This pure PDMS interface ensures superior bonding and structural integrity for the membrane attachment. Once the PDMS bulk, which consists of a thin layer of PDMS with microcavities and a thick layer of PDMS embedded with magnetic particles, was bonded to the 25 µm thick PDMS membrane, the microrobots were cut from the bulk PDMS using a 1 mm biopsy punch. Finally, the cut magneto-acoustic microrobots were held vertically on a cylindrical permanent magnet capable of generating high magnetic fields and subsequently magnetized. The cut membrane-based magneto-acoustic microrobot is later placed in a water chamber filled with PBS water mixed with non-ionic surfactant.

To maximize propulsion efficiency, the piezo transducer was driven at the membrane's resonant frequency by 140 volts peak to peak sinusoidal signal. The proposed acoustic microrobot utilizes a flexible PDMS membrane to convert the external acoustic energy into propulsive thrust, enabling motion. Actuation at this high voltage level not only ensures significant thrust generation but also demonstrates the structural resilience of the proposed design, suggesting the potential to withstand even higher acoustic power densities without compromising the integrity of the membrane-body interface.

Locomotion direction is governed by external magnetic fields. When the membrane-based magneto-acoustic microrobot is subjected to an external uniform magnetic field, the axially magnetized composite part of the acoustic microrobot body experiences a restorative torque ($T = m \times B$) that directs the alignment of the microrobot's internal magnetic moment with the external field vector. Due to the minimal frictional resistance within the water chamber, a relatively low magnetic field (~2 mT) generated by Helmholtz coils is sufficient to achieve high-precision orientation control.

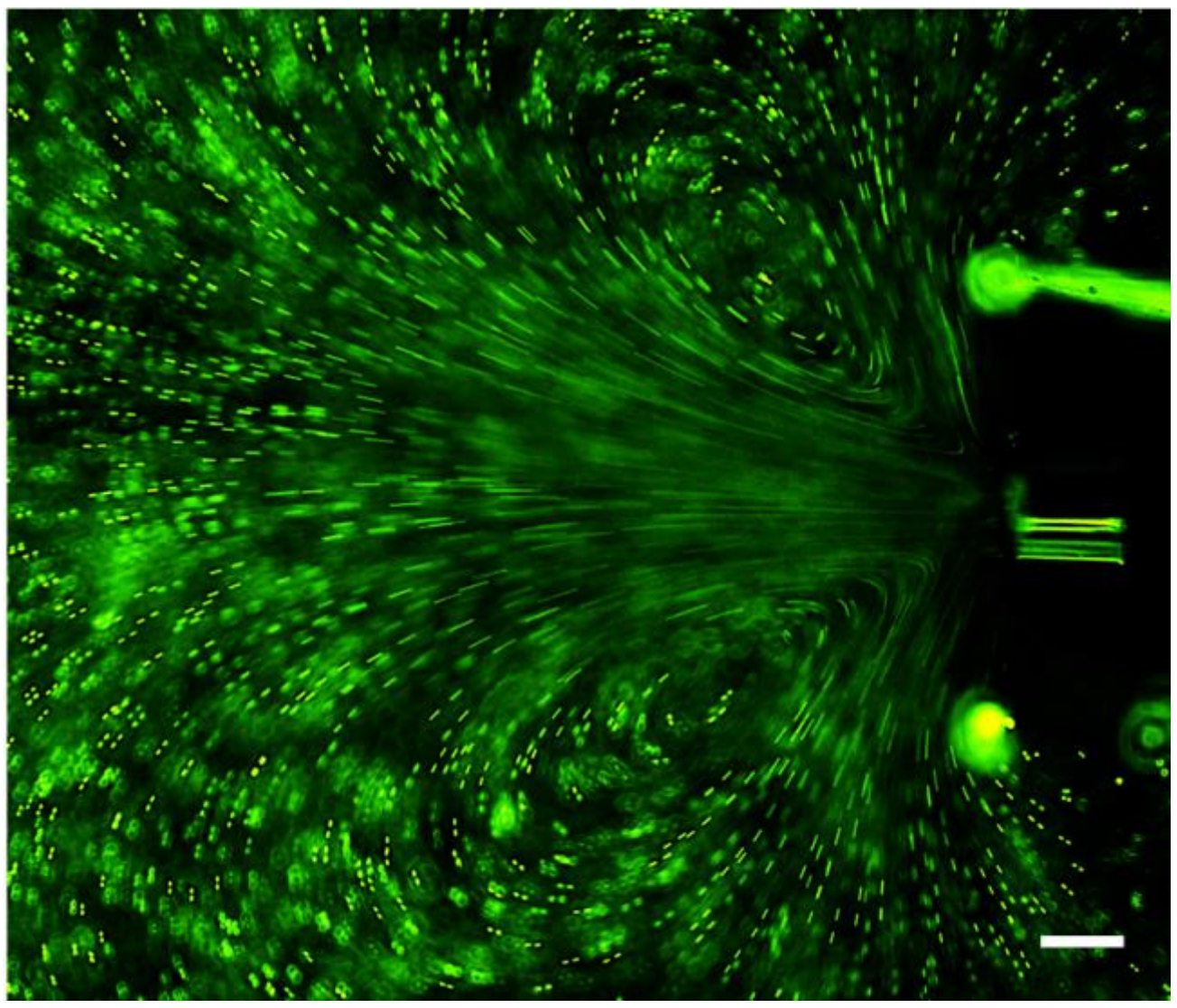

Figure 8. Microstreaming patterns of tracer particles generated by a scaled-down membrane-based acoustic actuator with a 120 µm x 120 µm diameter and depth, respectively. Scale bar, 100 µm.

### *D. Scaling Down Membrane-based Acoustic Microrobots*

To evaluate the scalability and functional viability of the proposed design at smaller length scales, a miniaturized prototype was fabricated (Figure 8). The molds were patterned using UV mask photolithography with a negative photoresist. The resulting architecture features microcavities with a diameter and depth of 120 µm, encapsulated by a bonded PDMS membrane approximately 8 µm in thickness.

Experimental observations of the acoustic streaming flow patterns (Figure 8) remained consistent with previous 350 µm diameter to 500 µm length microcavity counterparts, confirming that the thrust-generation mechanism scales effectively down to the 100 µm regime. As expected from acoustic scaling theory, the natural frequency of these miniaturized robots shifted significantly higher, with an observed resonance of approximately 60 kHz. This shift is attributed to the increased stiffness-to-mass ratio of the membrane.

## IV. Conclusion

In this paper, we present a novel method for generating acoustic propulsion that enables the design of durable, consistent, and robust acoustic microrobots. Experimental validation of the system's longevity, as well as the relationship between voltage and velocity, proves that the proposed method is highly robust and suitable for tasks requiring stable performance. Furthermore, we experimentally demonstrate a proof-of-concept for a navigable, membrane-based acoustic microrobot featuring embedded magnetic particles. While our current prototypes feature a single membrane-based propulsion unit, the stability of this method allows for the future development of multi-acoustic unit microrobots. Ultimately, optimizing hydrodynamic design will improve travel efficiency, though realizing membrane-based acoustic microrobots at below 50 µm necessitates further innovation in manufacturing methods.

## Acknowledgment

The authors would like to thank Direnc Akyildiz, Youyi Zhou, Alyssah Harnden, Poyraz Efe Olgun and Xiangyi Tan for their valuable support, technical assistance, and insightful discussions throughout this work. This work was supported by the National Science Foundation (NSF) Foundational Research in Robotics program (FRR- 2505842).